\documentclass{article}

\usepackage[final]{neurips_2024_ml4ps}

\usepackage[utf8]{inputenc} 
\usepackage[T1]{fontenc}    
\usepackage{hyperref}       
\usepackage{enumitem}
\usepackage{url}            
\usepackage{booktabs}       
\usepackage{nicefrac}       
\usepackage{microtype}      
\usepackage{xcolor}         
\usepackage{natbib}
\usepackage{tabularx,ragged2e,booktabs,caption}
\usepackage{subfigure}
\usepackage{multicol,tabularx,capt-of}
\usepackage{hhline}
\usepackage{wrapfig}
\usepackage{multirow}
\usepackage{multicol}
\usepackage{bm}
\usepackage{nomencl} 
\usepackage{algorithm}
\usepackage{algorithmic}
\usepackage{subcaption}
\usepackage{mathtools}
\usepackage{setspace}
\usepackage{natbib}

\title{Geometry-aware PINNs for Turbulent Flow Prediction}

\author{%
  Shinjan Ghosh, Julian Busch, Georgia Olympia Brikis, Biswadip Dey \\
  Siemens Corporation, Technology\\
  Princeton, NJ 08540 \\
}

\begin{document}

\maketitle

\begin{abstract}
Design exploration or optimization using computational fluid dynamics (CFD) is commonly used in the industry. Geometric variation is a key component of such design problems, especially in turbulent flow scenarios, which involves running costly simulations at every design iteration. While parametric RANS-PINN type approaches have been proven to make effective turbulent surrogates, as a means of predicting unknown Reynolds number flows for a given geometry at near real-time, geometry aware physics informed surrogates with the ability to predict varying geometries are a relatively less studied topic. A novel geometry aware parametric PINN surrogate model has been created, which can predict flow fields for NACA 4 digit airfoils in turbulent conditions, for unseen shapes as well as inlet flow conditions. 
A local+global approach for embedding has been proposed, where known global design parameters for an airfoil as well as local SDF values can be used as inputs to the model along with velocity inlet/Reynolds number ($\mathcal{R}_e$) to predict the flow fields. A RANS formulation of the Navier-Stokes equations with a 2-equation k-epsilon turbulence model has been used for the PDE losses, in addition to limited CFD data from 8 different NACA airfoils for training. The models have then been validated with unknown NACA airfoils at unseen Reynolds numbers.

\end{abstract}
\section{Introduction}
Over the last decade, deep learning has emerged as a vital tool for accelerating CFD simulations, particularly in applications where traditional solvers are computationally expensive or time-intensive \citep{vinuesa2022enhancing, warey2020data, zhang2022demystifying}. Several approaches have been developed, with some integrating neural networks to predict residuals or refine turbulence models within the framework of conventional solvers, effectively merging data-driven techniques with established numerical methods \citep{hsieh2019learning, doi:10.1073/pnas.2101784118}. Other strategies aim to replace the CFD solvers entirely by learning the flow fields directly through models that utilize convolutional neural networks (CNNs) or graph neural networks (GNNs). CNNs effectively capture the spatial features of the flow on regular grids \citep{hennigh2017lat, CHEN2023105707}, whereas GNNs excel at handling complex geometries and mesh representations, providing flexible and mesh-independent predictions \citep{jiang2020meshfreeflownet, wang2020towards}. Additionally, Physics-Informed Neural Networks (PINNs) have been introduced to embed governing partial differential equations (PDEs) into the loss function, ensuring that the models comply with the fundamental physical laws, such as the Navier-Stokes equations \citep{raissi2019physics, dwivedi2019distributed, lu2019deepxde, nabian2019deep, zhang2020frequencycompensated}. This physics-based regularization allows PINNs to predict the spatiotemporal evolution of flow variables while maintaining consistency with the underlying conservation laws, making them especially useful when dealing with sparse or noisy data. By incorporating physical constraints, PINNs maintain a balance between data-driven model accuracy and the theoretical rigor of traditional approaches.

Despite advances in Physics-Informed Neural Networks (PINNs), their use in RANS-based turbulence modeling, particularly for predicting flow at unseen Reynolds numbers, remains underexplored \citep{en16041755}. Previous studies often rely on the Reynolds-stress formulation, which demands high-fidelity data from costly simulations such as Direct Numerical Simulations, Large Eddy Simulations, or high-resolution flow measurements \citep{eivazi2022physics, HANRAHAN}. While this ensures accuracy, it limits scalability to practical applications. Recent efforts, such as \citet{patil}'s work with the Spalart-Allmaras (SA) model, have combined physics and data to offer a more computationally feasible approach. Similarly, \citet{pioch} evaluated models like $k$-$\omega$ and Reynolds-stress, using hybrid training strategies to reconstruct flow fields for fixed Reynolds numbers. \citet{ghosh2023_RANSpinn} applied the $k-\epsilon$ model with PINNs to predict turbulent flow for previously unseen Reynolds numbers, but their work did not address the influence of geometry which plays a critical role in design exploration and optimization. In this work, we bridge this gap by using Signed Distance Functions (SDFs) to incorporate geometry into PINN-based RANS modeling. SDFs are widely used in fields such as computer vision, graphics, and surrogate modeling \citep{10.1145/2939672.2939738, NEURIPS2023_70518ea4}. By embedding SDFs into the PINN framework, we aim to enhance the prediction accuracy over complex geometries and generalize flow reconstruction to novel geometries of the solution domain.

While a small body of work has investigated PINNs capable of accommodating geometry variations \cite{oldenburg2022geometry}, the use of PINNs to predict flow in novel geometries remains a less explored area. In this work, expand the scope of RANS-PINNs that combines PINN-based surrogates with RANS based turbulent modeling approaches and enable it predict flow and turbulent variables over new geometries. In particular, we use an SDF based geometric embedding along with a set of physically interpretable shape parameters to predict flow fields over any NACA 4-digit airfoil geometry and for any value of the inlet velocity. Key contributions can be listed as:
\begin{itemize}[noitemsep,topsep=-0.5em,leftmargin=*]
\item Accurate prediction of velocity and pressure fields for any Reynolds number($\mathcal{R}_e$) (within the range of prediction) over any given shape of a NACA 4-digit airfoil.
\item Incorporation of $k$-$\epsilon$ turbulence model and RANS based conservation equations to ensure data efficiency and improved generalization.
\item Usage of warm-start based training phase to ensure convergence when dealing with multiple PDE-based loss terms along with limited and shape varying CFD data. 
\item Evaluation of model performance an ablation study which takes into account local and global geometry information in form of SDF fields, and NACA 4-digit shape parameters.
\end{itemize}

\section{RANS-PINN Architecture and Training}
\label{gen_inst}

The constant density fluid flow is governed by the continuity equation (to \textit{conserve mass}) and Navier-Stokes equation (to \textit{conserve momentum}). A 2-equation eddy viscosity model describes the closure terms associated with the RANS formulation that is needed to capture the effect of turbulence. In this work, a standard $k$-$\epsilon$ turbulence mode is used. Symbols $U$ and $p$ denote the flow velocity ($x$ component $u$ and $y$ component $v$) and pressure, respectively. continuity and Navier-Stokes equations, at a density of  $1$ $kg/m^3$ can be expressed as:
\begin{align}
\textrm{\textbf{Continuity:}}\quad & 
\nabla(U) = 0
\\
\textrm{\textbf{Navier-Stokes:}}\quad &
(U\cdot\nabla)U + \nabla p-  \mu_{eff}\nabla^{2} U = 0,
\end{align}
where $\nabla$ denotes the vector differential operator, and $\mu_{eff}:= \mu + \mu_t= \mu + 0.09k^2/ \epsilon$ represents the effective viscosity, i.e., the sum of molecular viscosity ($\mu$) and turbulent viscosity ($\mu_t$). In addition, the $k$-$\epsilon$ turbulence model can be expressed as:
\begin{align}
\textrm{\textbf{$k$ - equation:}}\quad &
\nabla ( U k)
=
\nabla\left[\left( \mu+\frac{\mu_{t}}{\sigma_{k}} \right)\nabla k\right] + P_{k} - \epsilon
\\ 
\textrm{\textbf{$\epsilon$ - equation:}}\quad &
\nabla ( U \epsilon)
= 
\nabla\left[\left(\mu+\frac{\mu_{t}}{\sigma_{\epsilon}}\right)\nabla\epsilon\right] + (C_{1}P_{k}+C_{2}\epsilon)\frac{\epsilon}{k}
\end{align}
where, $C_1 = 1.44$, $C_2 = 1.92$, $\sigma_k = 1$, and $\sigma_{\epsilon} = 1.3$ are empirical model constants. Where, $P_k$ is the production term. The airfoil surface boundary conditions can be characterized as:

\begin{align}
\textrm{\textbf{Airfoil Surface Boundary Condition: }} U = 0, \textrm{ when } SDF = 0. 
\end{align}

\begin{figure}[t!]
\centering
\vspace{-1.20em}
\includegraphics[width=0.7\linewidth]{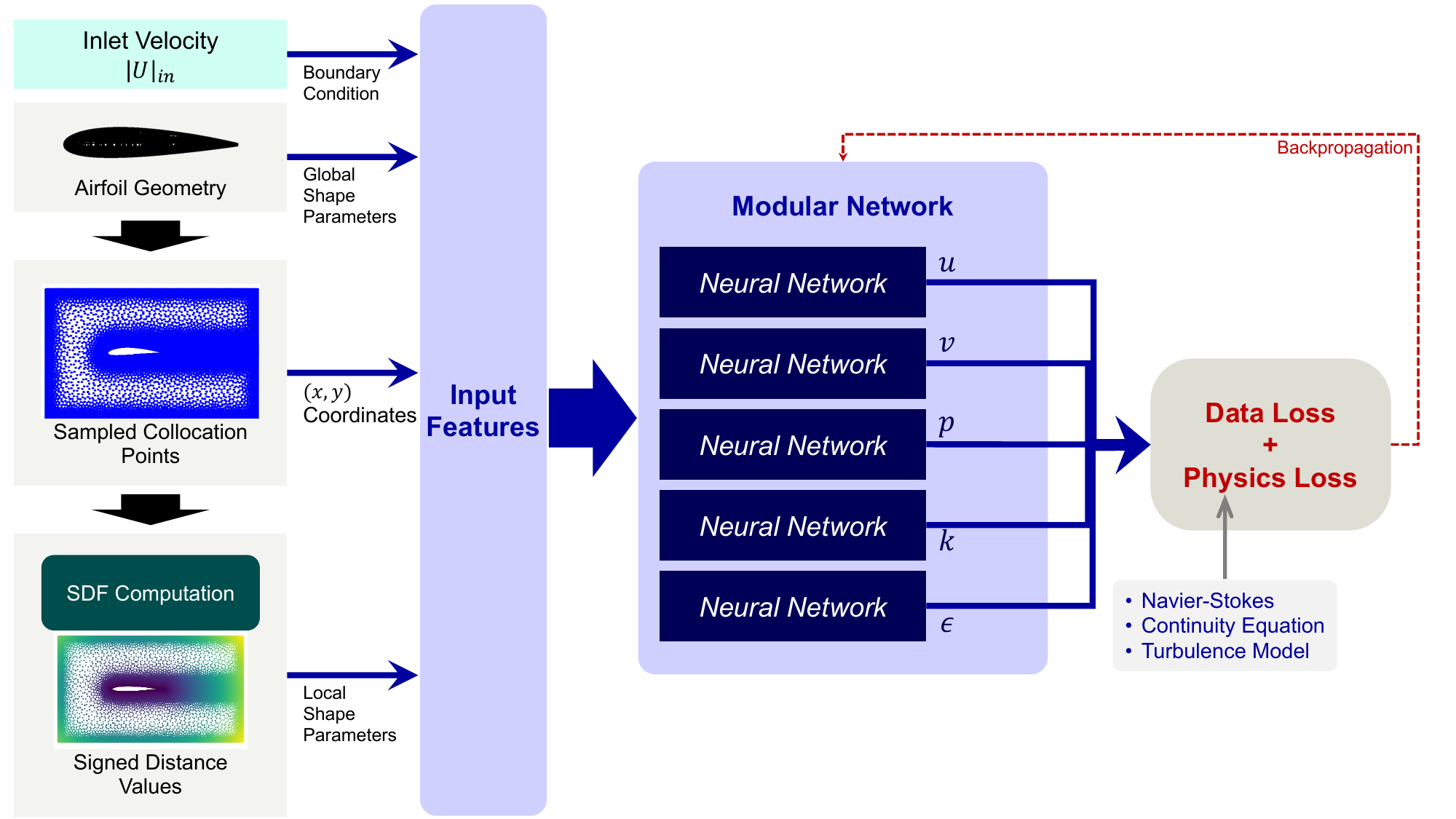}
\caption{\small{Architecture Diagram.}}
\label{airfoil_val}
\end{figure}

A CFD model with wake refinement and prism layers, with all wall $y^{+}$ has been created in STAR-CCM+ for each NACA airfoil. A velocity inlet boundary condition along with pressure outlets, no-slip walls and free stream side walls have been considered. A data set comprising of 8 NACA airfoils with, six inlet velocities($|U|_{in}$) randomly generated have been selected from the range of  2 m/s to 7 m/s, which corresponds to a Reynolds number ($\mathcal{R}_e=\rho|U|_{in}L/\mu$, where L=airfoil chord length) range of 200k to 700k to make a training set. For the first model (L), an SDF field has been calculated to embed the distance of each domain point from the geometry, there creating an invariant for each shape (Figure \ref{airfoil_val}).

In addition to this, the 4 digits from each NACA airfoil have been used as a design input for a second model (L+G). By using the inlet velocity as an additional input to the PINN, the surrogate has been parameterized to $\mathcal{R}_e$ variation. However, in case of a lack of knowledge of such design parameters for an unknown geometry, it has been shown that flow prediction is still possible with only the SDF fields. The performances have been compared against an ablation model (G) which has only the design parameters as an input in addition to the inlet velocity. A warm start approach is employed, where only the data losses are used to pre-train the PINN before introducing the PDE losses,  similar to \cite{ghosh2023_RANSpinn}, in order to avoid convergence issues with a set of complex governing PDEs. 


 \section{Results and Discussions}
Various turbulent flow features can be observed in a flow over an airfoil case. Due to acceleration of flow over the top surface, a pressure differential is formed between surfaces, which then results in lift. There is a stagnation zone formed at the nose, and a separation wake at the tail region. These 
\begin{figure}[h]
\centering
\vspace{-1.20em}
\includegraphics[width=0.7\linewidth]{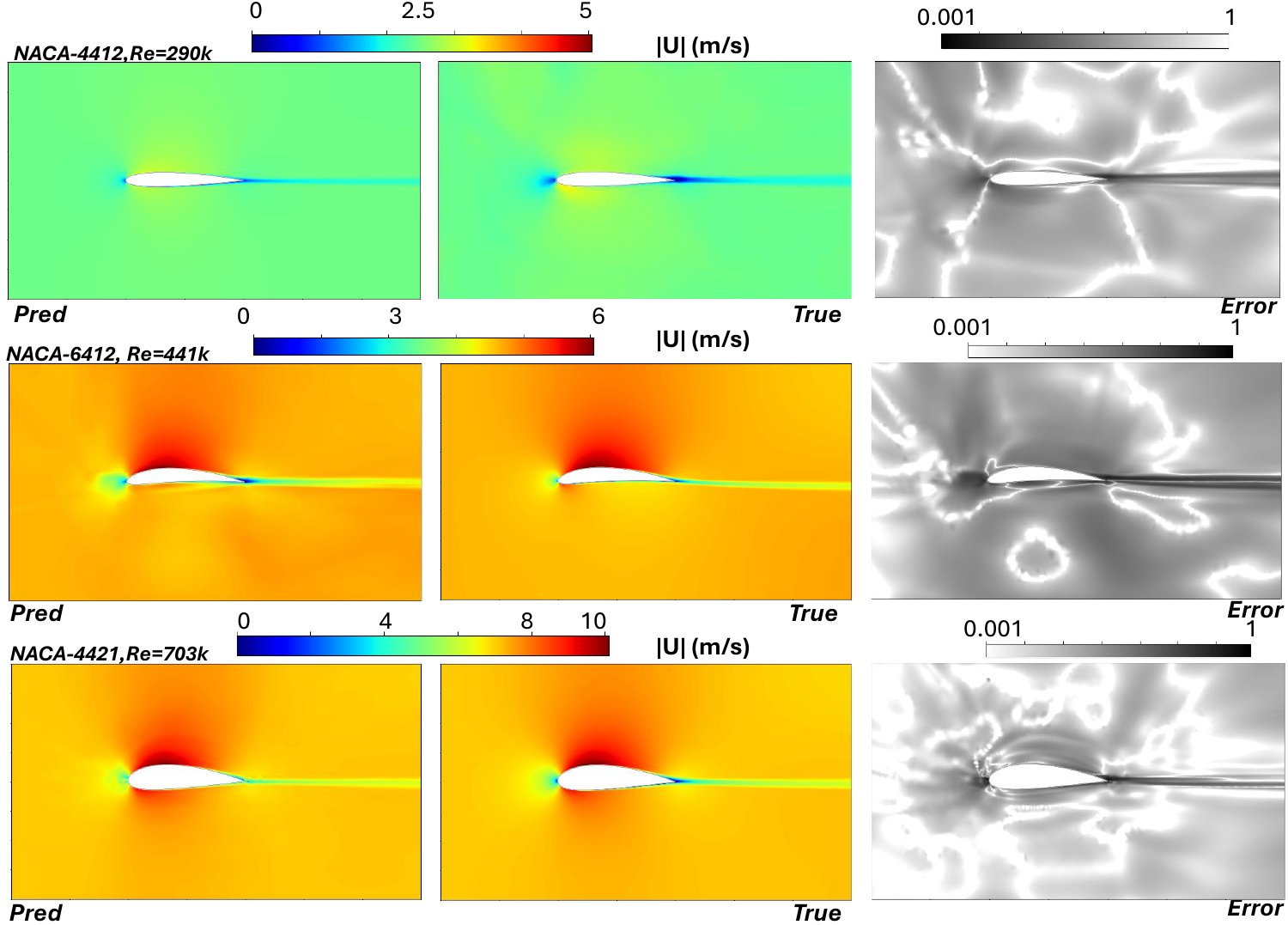}

\caption{\small{Velocity predictions for various NACA airfoils at different $\mathcal{R}e$.}}
\label{val_vel}
\end{figure}
features are well captured by the L model for all 3 unseen NACA airfoil geometries (Fig. \ref{val_vel}). The pressure plots (Fig. \ref{val_p}) show a peak at the nose, representing stagnation and a large low magnitude zone on the upper surface representing flow acceleration and pressure differential between the two surfaces. A multitude of shapes and inlet velocities have been studied in table 1, with an additional deeper dive into details in table 2. Table 1 shows the mean and median for velocity and pressure, which gives an idea about the overall distribution of errors inside the computational domain. As a high error is observed in near-wall regions, the overall mean can be skewed. 

The error for velocity is calculated as :
\begin{align}
    Error=(Pred-Ground Truth)/ Vel_{in}),
\end{align}
whereas, the error for pressure is calculated as:
\begin{align}
    Error=(Pred-Ground Truth)/ (0.5Vel_{in}^2).
\end{align}

Error plots in grey-scale also represent good match between the predicted and true cases, except narrow regions of high error at the walls, where the thin turbulent boundary layer is formed. Error magnitudes for predictions are observed in Table 2, where three different model types are evaluated at near and far zones. 

It is observed that the L model with the SDF input performs the best overall and the near zones, whereas the G model with only the design parameters as input is seen to perform worse overall and do better at the outer zone. This can be attributed to lack of local 
\begin{figure}
\centering
\includegraphics[width=.7\linewidth]{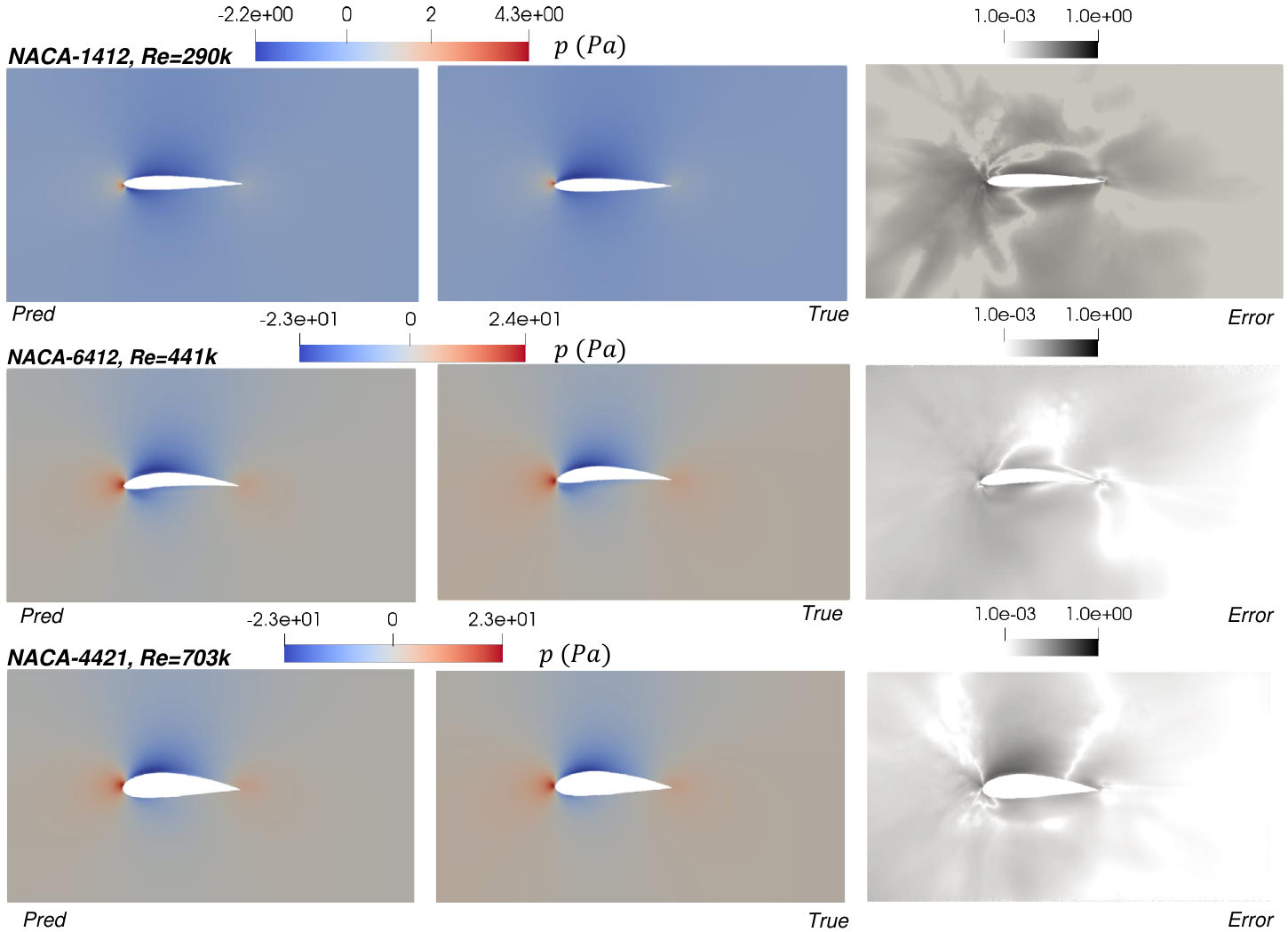}

\caption{\small{Pressure predictions for various NACA airfoils at different $\mathcal{R}e$.}}
\label{val_p}
\end{figure}
\begin{figure}[b]
\vspace{-1.25em}
\centering
\includegraphics[width=0.75\textwidth]{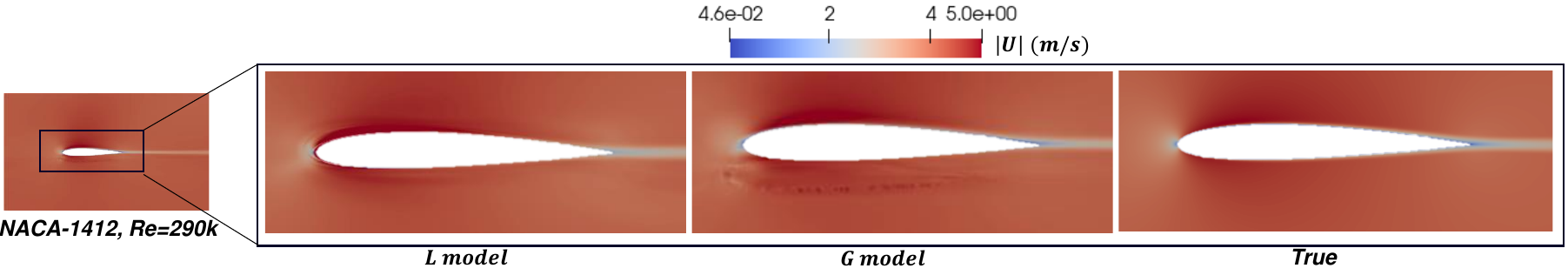}
\vspace{-0.5em}
\caption{\small{Prediction comparison of speed for L model and G model.  }}
\label{val_abl}
\vspace{-2em}
\end{figure}
information which the SDF provides in embedding the knowledge of geometric boundaries. Figure \ref{val_abl} shows the formation of phantom airfoil like shapes from the G model predictions. It is clear that this affects the performance in the near flow-field region (seen in the small rectangle on Figure 4). A third model (L+G) has then been created where both the SDF and the design parameters have been provided as input, to obtain a local and global understanding of the flow. It can be seen that the L+G model improves upon the performance of the G model, with slightly inferior performance in the near-field zone compared to the L model,and  performs better in the far field. Overall, the L+G model comes close to the metrics of the L model and even betters it in one instance(NACA-6414). This suggests, further investigation with more complex architectures accounting for local and global geometric information can possibly improve prediction further.
\subsection{NACA 4 digit airfoils and geometry embedding}

The NACA 4-digit airfoils are represented by the 3 parameters, which feature in the surface defining equations. Each unique value of a parameter can create a new airfoil. The naming convention can be defined by:\\
\indent
\textbf{NACA-MPXX}, where $M$ = max camber, $P$ = position of max camber and $XX$= thickness.

The G model uses these parameters as an addition to the inputs of a standard PINN, in order to embed geometry into the parametric RANS-PINN from previous literature \citep{ghosh2023_RANSpinn}. The L model ignores the parameters and ingests the computational domain in form of an SDF field in addition to the coordinates. The SDF in combination with the coordinates can help store information about the relative position of the point cloud points wrt the airfoil surface. The L+G model, however uses the combined information of the local SDF parameter in addition to the design parameters of the 4-digit airfoil to combine the best features of both approaches. 

\begin{table}
\centering
\caption{\small{Normalized validation $\%$ errors(mean and median) for pressure and velocity magnitude }}
\begin{tabular}{c|cccc}
\toprule
Validation &  \multicolumn{2}{c}{Velocity Magnitude ($\|U\|$)} & \multicolumn{2}{c}{Pressure ($p$)}
\\
Case & Mean Error & Median Error & Mean Error & Median Error
\\
\midrule  
NACA-1412, $\mathcal{R}_e=290\textrm{k}$ &
6.00 & 1.90 & 3.30 & 1.80 
\\
\midrule  
NACA-2115, $\mathcal{R}_e=410\textrm{k}$ &
3.42 & 0.64 & 2.54 & 0.44
\\   
\midrule  
NACA-2408, $\mathcal{R}_e=210\textrm{k}$ &
4.66 & 1.46 & 3.83 & 1.54
\\
\midrule  
NACA-1412, $\mathcal{R}_e=389\textrm{k}$ &
5.55 & 1.90 & 1.75 & 0.29
\\
\midrule  
NACA-4421, $\mathcal{R}_e=703\textrm{k}$ &
3.51 & 0.47 & 2.21 & 2.86
\\
\midrule  
NACA-6412, $\mathcal{R}_e=441\textrm{k}$ &
6.02 & 2.35 & 5.08 & 2.28
\\
\bottomrule
\end{tabular}
\label{Table_1_allerr}

\end{table}

\section{Conclusions}
A parametric PINN is developed to predict flow fields at unseen $\mathcal{R}_e$ and unseen airfoil geometries for incompressible turbulent flows. The RANS formulation, commonly used in industrial applications due to its computational efficiency, can become expensive for repeated simulations in the case of design exploration.  The proposed PINN surrogate leverages an SDF embedding of airfoil geometries, in addition to physics losses from RANS formulation with 2-equation turbulent models. Parameterized with respect to varying geometries and inlet velocities, the current PINN surrogate is able to predict flows over unknown for various NACA airfoils at different $\mathcal{R}_e$ and geometries. 
\begin{wraptable}[14]{r}{0.75\linewidth}
\vspace{-0.75em}
\caption{\small{Normalized validation errors for velocity magnitude }}
\vspace{-0.75em}
\begin{tabular}{lccccr}
\toprule
Validation case & L model & G model  & L+G model &
\\
   \midrule
   NACA 1412, $\mathcal{R}_e$=290k (Near)
 & \textbf{5.7}& 9.6  & 7.8& \\
 NACA 4421, $\mathcal{R}_e$=703k (Near)
   & \textbf{5.2 }& 13.8& 7.1& \\
 NACA 6412, $\mathcal{R}_e$=441k (Near)
  & \textbf{4.9} & 7.6 & 5.4 &\\
  \midrule
 NACA 1412, $\mathcal{R}_e$=290k (Far)
 & 3.2 & \textbf{2.7} & 3.0& \\ 
NACA 4421, $\mathcal{R}_e$=703k (Far)
   & 2.5 & 4.4 & \textbf{2.5}& \\
NACA 6412, $\mathcal{R}_e$=441k (Far)
  & 4.2 & 3.0  & \textbf{2.9}  &\\
\midrule
 NACA 1412, $\mathcal{R}_e$=290k
   & \textbf{3.5}& 5.0  & 4.9& \\
 NACA 4421, $\mathcal{R}_e$=703k 
   & \textbf{3.4} &6.7 & 4.1& \\
 NACA 6412, $\mathcal{R}_e$=441k
  & 4.4 & 4.6 & \textbf{3.7} &\\
\bottomrule
\end{tabular}
\vspace{0.2em}
\end{wraptable}
A comparison has been made with two other models which don't use SDF(G) and use both design parameters as well as SDF(L+G) representation of the geometry. The first model remains the best performing one, while more investigation is necessary to analyze how a local+global geometry aware model can be further optimized.
\newpage
\small
\bibliography{ML4PS_paper.bib}
\bibliographystyle{abbrvnat}

\end{document}